\documentclass[runningheads]{llncs}
\usepackage{graphicx}
\usepackage{amsmath,amssymb} 
\usepackage{svg}
\usepackage{color}

\begin{document}
\pagestyle{headings}
\mainmatter

\def\ACCV20SubNumber{924}  

\title{Object Detection-Based Variable Quantization Processing} 
\titlerunning{Object Detection-Based Variable Quantization Processing}
\authorrunning{L. Liu, H. Qi}
\renewcommand{\thefootnote}{\fnsymbol{footnote}}
\newcommand*\samethanks[1][\value{footnote}]{\footnotemark[#1]}
\author{Likun Liu\thanks{Both authors contributed equally to this research.}
\and Hua Qi\samethanks}
\institute{Kyushu University}
\maketitle

\begin{abstract}
  In this paper, we propose a preprocessing method for conventional image and video encoders that can make these existing encoders content-aware. By going through our process, a higher quality parameter could be set on a traditional encoder without increasing the output size. A still frame or an image will firstly go through an object detector. Either the properties of the detection result will decide the parameters of the following procedures, or the system will be bypassed if no object is detected in the given frame. The processing method utilizes an adaptive quantization process to determine the portion of data to be dropped. This method is primarily based on the JPEG compression theory and is optimum for JPEG-based encoders such as JPEG encoders and the Motion JPEG encoders. However, other DCT-based encoders like MPEG part 2, H.264, etc. can also benefit from this method. In the experiments, we compare the MS-SSIM under the same bitrate as well as similar MS-SSIM but enhanced bitrate. As this method is based on human perception, even with similar MS-SSIM, the overall watching experience will be better than the direct encoded ones.

\end{abstract}

\section{Introduction}

Presently, video content occupies more than 80\% of global internet traffic \cite{Cisco}. With services delivered in video format grows exponentially, this percentage is expected to be higher in the foreseeable future. However, not all network bandwidths in the world have grown along with this trend. Thus, better encoders are needed to deliver these contents broader.

Over the past few decades, numerous image and video encoders \cite{ISO10918,ISO14496,ISO15948,Wiegand2003,Sullivan2012} have emerged to suits the needs. Consumer video services are primarily in lossy format utilizing lossy video codecs to save spaces and bandwidth. These lossy encoding methods are mainly based on the theory of human perception and heavily rely on processes like quantization to reduce the data size. For a single image or video, the quantization matrix is invariable since the same matrix is required to recover the image during the decoding stage. These kinds of compressing strategies globally apply the same process to every block in the frame regardless of the actual content.

Recently, serval deep neural network (DNN) based auto-encoder for image compression \cite{Toderici2015,Balle2016,Toderici,Balle2018,Johnston2018,Theis2017,Li2018,Rippel2017,Agustsson2019} has achieved relatively high performance in comparison with traditional methods. However, an inconvenient fact is that these encoders would consume a massive amount of computing power to achieve their goal. Such methods may be a viable option for service providers who possess server clusters. Nevertheless, for applications like field live streaming, the bandwidth and the computing power may be heavily shackled as a result of the stringent on-site situation. 

Another technology that rises along with the mass utilization of GPU power is neural network-based object detection algorithms \cite{lin2013network,Krizhevsky2017,Szegedy2015,He2016}. They have achieved a relatively high precision in comparison with traditional recognition methods, and in the wake of recent optimization and miniaturization \cite{Redmon2016,Redmon2017,redmon2018yolov3,bochkovskiy2020yolov4}, object detection tasks can be done within a reasonable computing resource.

\begin{figure}
\centering
\includegraphics[width=120mm]{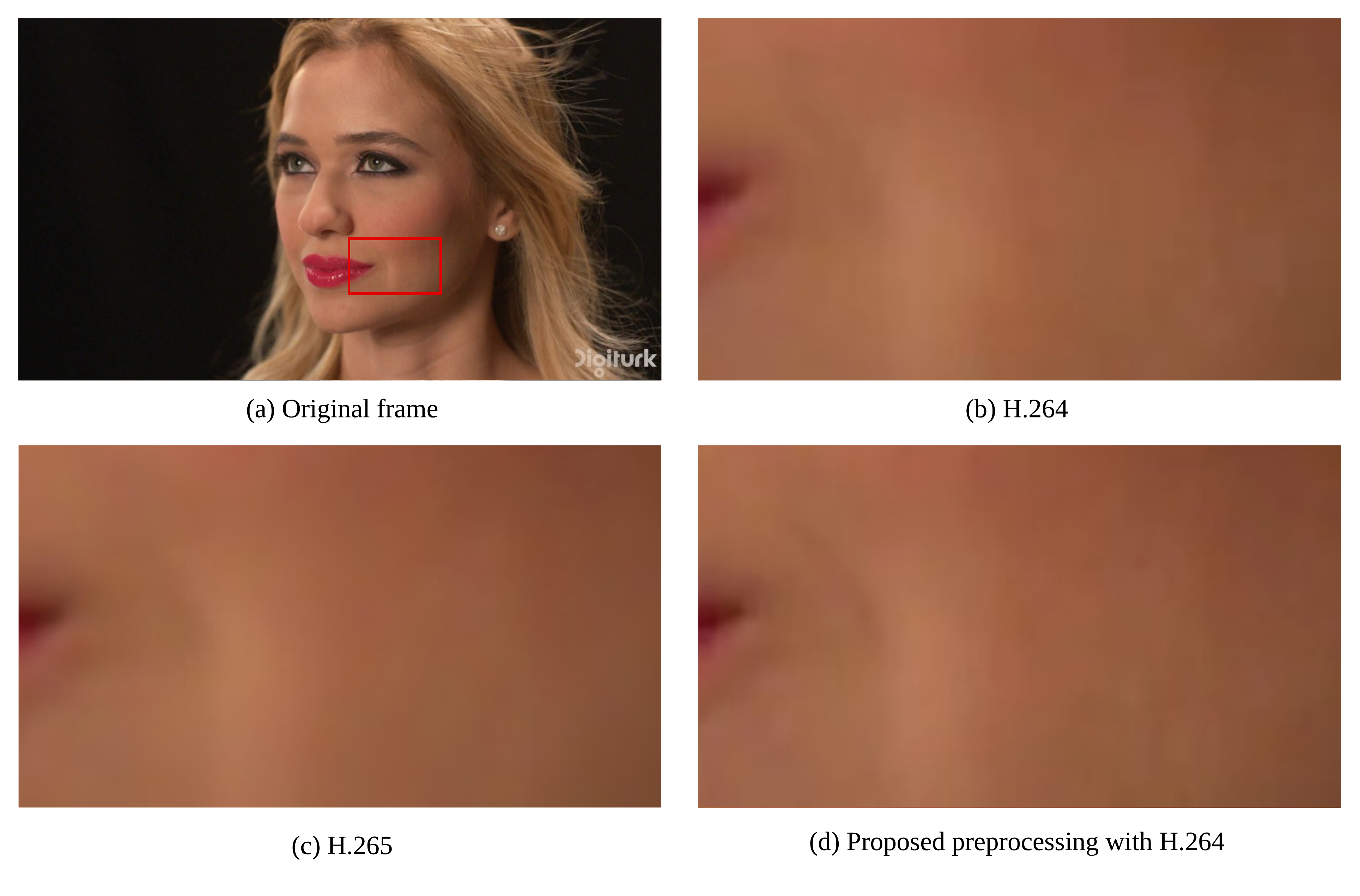}
\caption{
Visual comparison between video encoders. Picture (a) is the uncompressed frame. (b)-(d) are the enlarged detailed comparison between H.264, H.265, and H.264 with our proposed preprocessing method, respectively.}
\label{fig:demo}
\end{figure}

To the best of our knowledge, we present the first object detection based preprocessing method to make the encoders content-aware. Fig .~\ref{fig:demo} presents a visual comparison between the conventional encoders' direct encoding output and the encoding output with our preprocessing method. Noticing the subtle difference in the shadow area. With our preprocessing method, the encoders are able to preserve more details in region of interest. By keeping the detected objects in the scene untouched and process the remaining part with a relatively aggressive compressing approach, we were able to preserve more details of the object in the scene to enhance the quality of the video under the same bandwidth condition. In short, we have moved the resource from a trivial background to the main objects in the scene. The ascendancies of our proposed method are as follows: 
   \begin{itemize}
     \item When integrating with a DCT-based encoder (JPEG, H.264, etc.), the quantization matrix of the encoder is no longer invariable for the whole image in equivalent. As negligible parts of a frame have already processed with an aggressive quantization matrix, a relatively high-quality factor could be set within the same compression rate. 
     \item The object identification process was achieved with YOLOv4 \cite{bochkovskiy2020yolov4} networks, as this kind of object detection network is highly customizable in both network size and categories to detect, our method could adapt to different scenarios accordingly. 
     \item The aggressiveness of the quantization matrix is automatically varied according to the content of the frame. Experimental results will show that our approach will have a relatively high SSIM and SNR on other methods; also, the framerate would be higher under the same condition. 
     \item We have implemented all our processes parallelly on TensorFlow Framework and have achieved a remarkable 200 times acceleration rate over the sequential pure CPU version. The selection of the YOLOv4 detector allows the procedure to run at high speed even for a tiny GPU. 
   \end{itemize}


\section{Related Work}

\subsection{Object Detection}

Visual recognition task has been a significant research hot-spot recently. Numerous novel methods have been proposed during the past few years. These proposed methods \cite{lin2013network,Krizhevsky2017,Szegedy2015,He2016,touvron2020fixing} have reached an increasingly insane level of accuracy with the expansion of the neural network’s size in both layers and parameters. To the best of our knowledge, the world’s most accurate image recognition network at this time point is FixEfficientNet-L2 \cite{touvron2020fixing}, which has reached 98.7\% accuracy with 480 million parameters. However, these carefully designed high-accuracy networks are monumental and require a vast amount of computing resources on both the training and recognition process. This deficiency makes these nets unsuitable for video compression or live streaming in most of the scenarios since personal computers account for the vast majority of such applications. 

In contrast to these region proposal series of algorithms, a category of pure CNN based recognition algorithms stands out for its trade-off between accuracy and computing resources \cite{Redmon2016,Redmon2017,redmon2018yolov3,bochkovskiy2020yolov4,howard2017mobilenets}. The recent YOLOv4 \cite{bochkovskiy2020yolov4} can reach 65.7\% AP50 accuracy on the MS COCO dataset and running at ~65 FPS on a Tesla V100. This feature provides the possibility of utilizing this method within an affordable computing power range. Also, for tiny objects that are hard to be recognized by the YOLO detector, are likely to be the negligible objects in the scene in which insignificant to human perception.

\subsection{Image Compression}

A considerable amount of image compression algorithms have been proposed as the multi-media content gradually occupies the network traffic. Traditional lossy compression methods \cite{ISO10918,ISO15948} are consisting of carefully designed handcraft techniques. These techniques are a combination of human perception, signal processing, and experience. For instance, the most popular JPEG compression \cite{ISO10918} utilizes the YUV color encoding system since the human eyes are more sensitive to the luminance than the color. A frame is separated and quantified in Y, Cb, Cr channels individually, and each channel is quantified with a separate quantization parameter. Although a large number of quantization parameters can be chosen for different tasks, one apparent defect of this method is that they rely on user designation (quality factor) or frequency evaluation. This would cause severe detail loss under a limited bandwidth scenario. 

Another category of image compression method that rise along with the neural networks is DNN-based image compression \cite{Toderici2015,Balle2016,Toderici,Balle2018,Johnston2018,Theis2017,Li2018,Rippel2017,Agustsson2019,Ma2019}. Some of these methods utilized recurrent neural networks(RNNs) \cite{Toderici2015,Toderici,Johnston2018} to build a progressive image compression scheme while other methods \cite{Balle2016,Balle2018,Theis2017} exploits the power of the CNNs. These methods have achieved slightly higher performance than the traditional encoders in some applications. However, many of them are still suffering performance issues and cannot be applied to most of the scenarios. 
\subsection{Lossy Video Compression}

In the DCT based video encoding process,  the quantization process may slightly vary from the image compression. For instance, H.264 \cite{Wiegand2003} encoders utilize QP (Quantizer Parameters) for the quantization process. These QPs corresponding to a unique Qstep. In total, the H.264 encoder has 52 different Qstep values corresponds to 52 QPs. Similar to the JPEG quantization process, the luminance channel and the chromatic channels are treated with different QPs. In general, the Qsteps of the luminance channel ranges from 0 to 52, while the Qstep for the chromatic channels ranges from 0 to 39. However, they are still based on the same frequency theory, and our method will still function well on these encoders. 

The latest HEVC (H.265) \cite{Sullivan2012} encoding utilizes roughly the same encoding framework as the H.264. However, in almost every module, the HEVC added new encoding methods, including quadtree-based block division, inter-frame merge, AMVP technology, variable-size DCT, cabac, loop filtering, SAO, etc. In theory, with our proposed processing method, the variable-size DCT will further enhance the encoding efficiency of the encoder.

Both the AVC (H.264) and the HEVC (H.265) encoding supports lossless encoding, which will encode each frame as a lossless still frame. However, this encoding parameter will result in significantly large file size (usually more than 100 times bigger) and is seldomly used in practice. 

As the traditional video encoders are heavily modulized, DNN-based module enhancement and process optimization have been proposed. These proposals including intra prediction and residual coding \cite{Chen2018}, mode decision \cite{Liu2016}, entropy encoding \cite{Ma2019} and post processing \cite{Theis2017}. However, like DNN-based methods in image compression, these methods are either consume massive computing resources or are incompatible with traditional encoders and therefore require unique decoders to decode. 

\section{Proposed Method}
\subsection{Brief introduction of Image Compression}
In this section, we provide a brief introduction of the image compression method, specifically the JPEG’s quantization process. The conventional JPEG encoding procedure is shown as follows:

\begin{enumerate}
  \item Color space transformation and downsampling 
  
  JPEG utilizes YUV color space instead of a conventional RGB color encoding system. The input RGB image is firstly converted into YUV color space, in which we will explain the method in the next section. Afterward, if necessary, the converted image will go through the downsampling process in which the data in the U channel and V channel will be dropped. 
  
\item Block splitting and discrete cosine transform 

This process split each channel in the image into 8X8 macroblocks. If the image’s size does not satisfy the integer number of blocks, specific methods are exploited (fill with zeros or repeat edge pixels) to meet the requirements.   

Next, each macroblock will be converted into the frequency domain where low frequency in the top-left corner and high frequency in the bottom-right corner. 

\item Quantization

Human eyes are excellent in distinguishing low-frequency signals but may not be so sensitive regarding the high-frequency signal. Based on this fact, each macroblock is divided by an 8X8 quantization matrix, and each value will be rounded to the nearest integer to reduce the information contained in the image. The 8X8 quantization matrix is an experience-based matrix with generally low integer value on the top-left corner and high integer value on the bottom-right corner. According to the nature of DCT, this will result in low-frequency signal well preserved and high-frequency signal round to zero. 

\item Entropy encoding 

This step takes the result of quantization and performs lossless data compression. In JPEG compression, this step uses Huffman encoding \cite{Huffman1952} in the “zig-zag” order. Since most oblique data in the lower right corner is zero, this process could dramatically reduce the redundant data. 

\item Image reconstruction 

This procedure is essentially the reverse transformation of the above steps in reverse order. In the quantization phase, the result was rounded to the nearest integer and is irreversible, which made this compression method lossy compression. 
\end{enumerate}

\subsection{Overview of the Proposed Method}

\begin{figure}
\centering
\includegraphics[width=120mm]{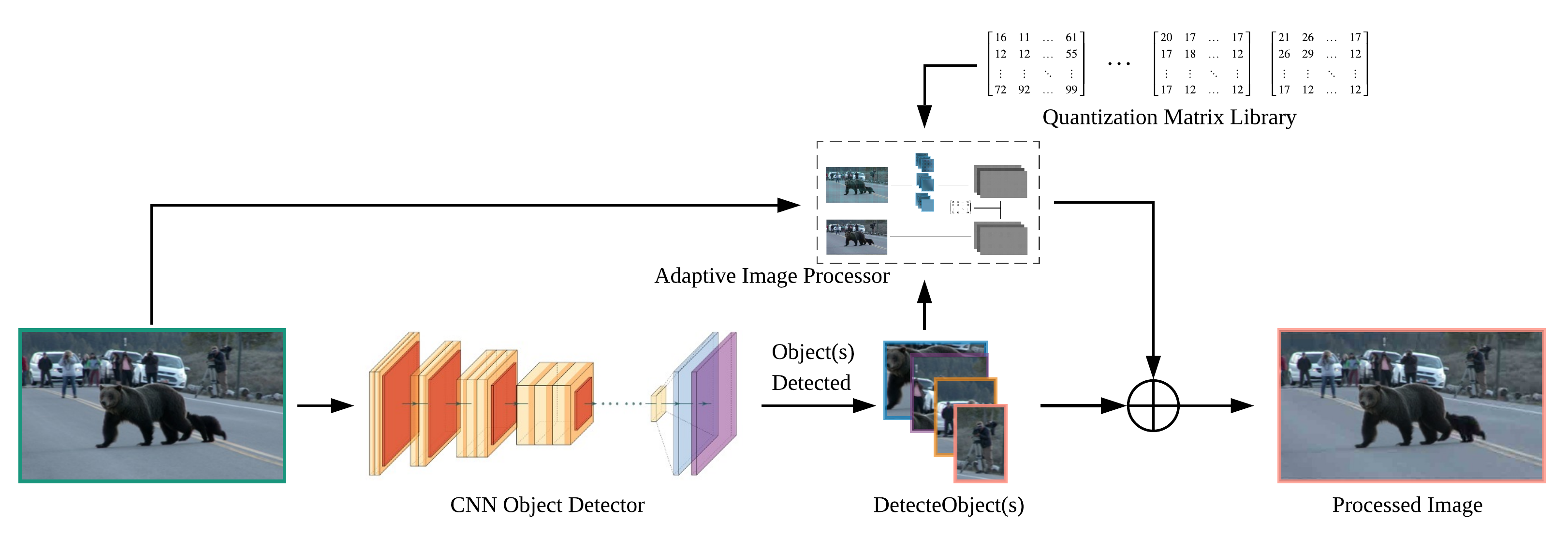}
\caption{
Architecture of the proposed preprocessing method}
\label{fig:architecture}
\end{figure}

Fig .~\ref{fig:architecture} presents a summarized diagram of our processing system. As this is a preprocessing method, the output is passed on to a conventional encoder. The procedure our proposed method is briefly summarized as follows:
\begin{itemize}

    \item \textbf{Object detection via YOLOv4: }Feed the frame or the still image directly into a YOLOv4 detector. Let \(O=\{o_1,o_2,\cdots,o_n\}\) denotes the collection of detected objects and the objects in the frame.  If no object is detected \(O=\emptyset\), the whole procedure will be bypassed, and no transformation will be applied to the image. If one or more object is detected and the confidence exceeds the threshold, the object’s coordinates and the corresponding image data will be saved.
    
    \item \textbf{Color space transformation and DC shifting:  }Let \(I\) denotes the given frame and \(I_R,I_G,I_B,I_Y,I_U,I_V\) denotes the corresponding channel of the frame. If an object is detected in the previous step, the whole image will be transformed into YUV color space. A simplified transformation equation in Rec. 601 \cite{BT601} represent in matrix form can be defined as:
    \begin{equation}
        \begin{bmatrix}
        I_Y\\
        I_U\\
        I_V\\
        \end{bmatrix}
        =
         \begin{bmatrix}
        66 & 129 & 25\\
        -38 & -74 & 112\\
        112 & -94 & -18\\
        \end{bmatrix}
        \begin{bmatrix}
        I_R\\
        I_G\\
        I_B\\
        \end{bmatrix}
    \end{equation}
    
    To reduce the fluctuation in absolute value, we have shifted the entire image down by 127, namely minus every pixel’s luminance value in each channel by 127.
    
    \begin{equation}
        I_{yuv\_shifted}=I_{yuv}-127*ones(I.shape)
    \end{equation}
    
    \item \textbf{Block splitting and Discrete Cosine Transform:   }Similar to the JPEG compression process, we split the shifted image into \(8\times8\) blocks:
    
    \begin{equation}
        B_{i,j}=I_{yuv\_shifted} [(8i:8i+7,8j:8j+7)]
    \end{equation}
    
    Where \(B_{i,j}\) represents the macroblocks of the shifted image \(I_{yuv\_shifted}\) and perform block-wise 2D discrete cosine transform. The equation which transforms the time-domain signal into the frequency domain with a step of eight is shown as bellow: 
    
    \begin{equation}
        F_{(u,v)}=\frac{1}{4}\alpha(u)\alpha(v)\sum_{x=0}^{7}\sum_{y=0}^{7}f_{(x,y)}\cos{\left[\frac{(2x+1)u\pi}{16}\right]}\cos{\left[\frac{(2y+1)v\pi}{16}\right]}
    \end{equation}
    
    \begin{equation}
        \alpha(u),\alpha(v)=
        \begin{cases}
            1/\sqrt{2},& \text{if } u,v = 0\\
            1,              & \text{otherwise}
        \end{cases}
    \end{equation}
    
    Where \(f_{x,y}\) denotes the pixel value in the \(8\times8\) macroblock \(B_{i,j}\) and \(F_{u,v}\) denotes the corresponding value in the frequency domain.
    
    In practice, we have composed a transformation matrix and utilize matrix multiplication to perform the DCT procedure.  Define an \(8\times8\) orthogonal matrix \(C\) where \(C^T\cdot C=E\) as:
    \begin{equation}
        C=\frac{1}{2}
        \begin{bmatrix}
            \sqrt{1/2} & \sqrt{1/2} & \cdots & \sqrt{1/2} \\
            \cos{\frac{\pi}{16}} & \cos{\frac{3\pi}{16}} & \cdots & \cos{\frac{15\pi}{16}} \\
            \vdots & \vdots & \ddots & \vdots \\
            \cos{\frac{7\pi}{16}} & \cos{\frac{21\pi}{16}} & \cdots & \cos{\frac{105\pi}{16}} \\
        \end{bmatrix}
    \end{equation}
    
    The DCT process can be simplified as:
    
    \begin{equation}
            F=C^T\cdot f \cdot C
    \end{equation}

    \item \textbf{Adaptive Quantization:   }In this stage, the previously processed blocks will be divided by a quantization matrix individually, and each point is rounded to its nearest integer. The quantization matrix used in this step is variable according to the frame contents and the respective channel. Then, the frequency domain matrices will be multiplied by the same quantization matrix to restore its data. Details of this process will be illustrated in the next section.
    
    \item \textbf{Inverse DCT and DC shifting: }The final step is to put the restored frequency matrix through a 2D inverse discrete cosine transformation, as shown below:
    
    \begin{equation}
        f_{(u,v)}=\frac{1}{4}\alpha(u)\alpha(v)\sum_{x=0}^{7}\sum_{y=0}^{7}F_{(x,y)}\cos{\left[\frac{(2x+1)u\pi}{16}\right]}\cos{\left[\frac{(2y+1)v\pi}{16}\right]}
    \end{equation}
    
    \begin{equation}
        \alpha(u),\alpha(v)=
        \begin{cases}
            1/\sqrt{2},& \text{if } u,v = 0\\
            1,              & \text{otherwise}
        \end{cases}
    \end{equation}
    
   Where \(f_{x,y}\) denotes the pixel value in the \(8\times8\) macroblock \(B_{i,j}\) and \(F_{u,v}\) denotes the corresponding value in the frequency domain. Similar to the DCT, the IDCT process can also be simplified using the same orthogonal matrix \(C\): 
   
       \begin{equation}
            f=C\cdot F \cdot C^T
        \end{equation}
        
        To restore the image, a DC value of 127 will be added back to the image. In step 4, the value of each point is rounded to its nearest integer, and there is a possibility that pure white and pure black pixels will be higher or lower than the tolerant range after shifting. To suppress the overflow, the entire matrix will be clipped within a 0-255 range. 
        
        \begin{equation}
            I_{yuv\_restored}=clip(I_{yuv\_quantified}+127\times ones(I.shape))
        \end{equation}
    
    \item \textbf{Object restoration:    }This step takes the processed image from the previous step and the saved image fragments from step 1 to synthesis the final output image. Each fragment is returned to its original position, respectively.
    
        \begin{equation}
            I_{processed}=I_{yuv\_restored}\oplus o_1 \oplus o_2\oplus \cdots \oplus o_n
        \end{equation}

    \item \textbf{(Optional) Color space transform:    }The restored image is in YUV color space, if required, it will be transformed back into RGB color space.
\end{itemize}


\subsection{Adaptive Quantization Process}

\begin{figure}
    \centering
    \includegraphics[width=120mm]{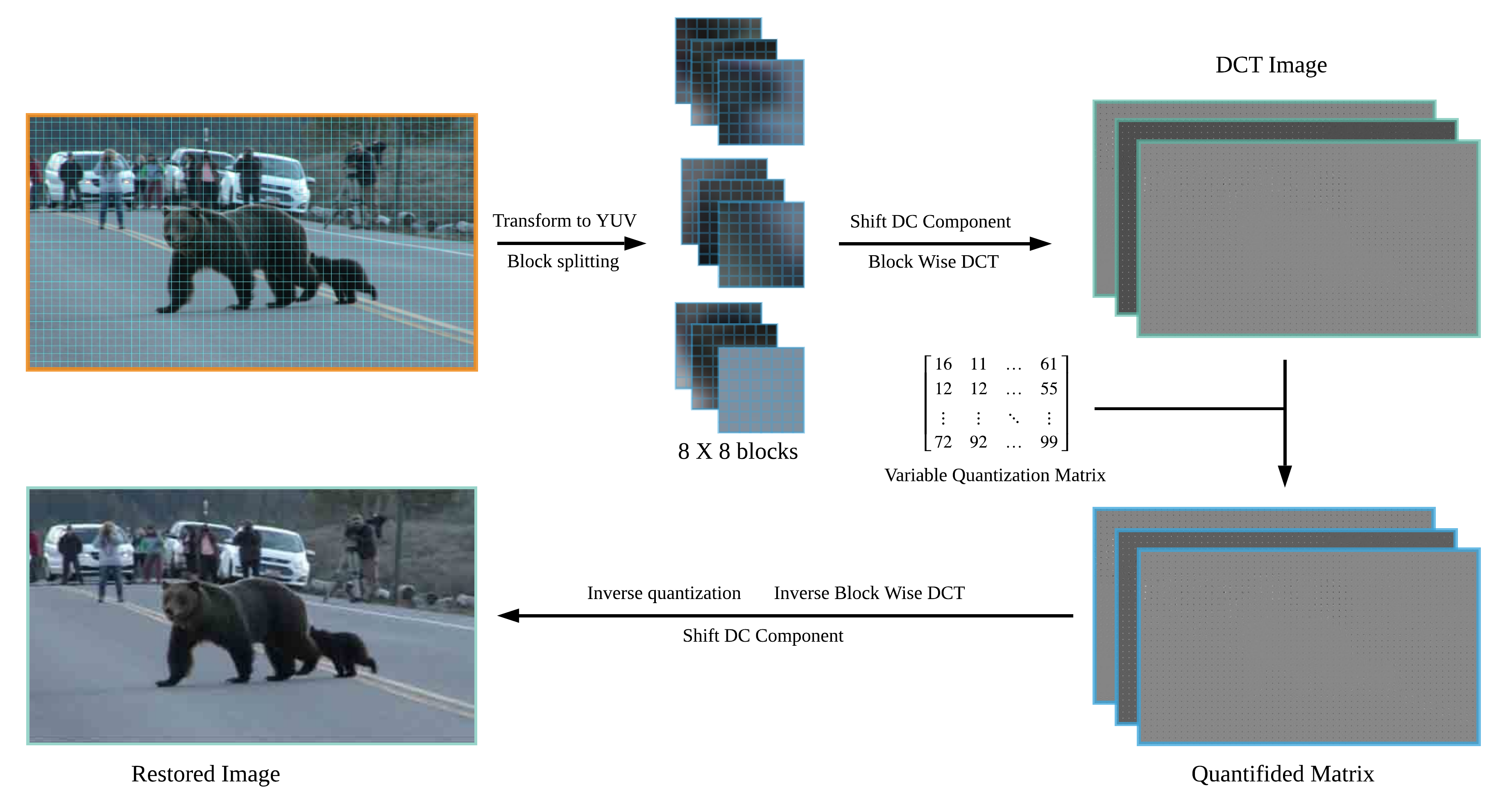}
    \caption{Adaptive Quantization Process}
    \label{fig:Quantization_Detail}
\end{figure}

We have pre-defined eight types of quantization matrices divided into two categories, denote as \(L_0,L_1,L_2,L_3\) and \(C_0,C_1,C_2,C_3\) respectively. The \(L_m\) matrices are mild matrices for luminance channel (Y channel) quantization while the \(C_n\)  matrices are relatively aggressive matrices for chromatic channels (U and V channel). The footnote 0-3 indicates the quality factor of each matrix. The detail of the process is shown in Fig .~\ref{fig:Quantization_Detail}.

 In step3, we obtained the frequency matrix of the macroblock \(B_{i,j}\) denote as \(F\) and \(F_Y,F_U,F_V\) for the respective channel. Each channel is divided by its corresponding quantization matrix and rounded to the nearest integer value:
\begin{equation}
    \begin{cases}
    F_{Y\_quantified(u,v)}=round(F_{Y(u,v})/L_{m(u,v)})\\
    F_{U\_quantified(u,v)}=round(F_{U(u,v})/C_{n(u,v)})\\
    F_{V\_quantified(u,v)}=round(F_{V(u,v})/C_{n(u,v)})\\
    \end{cases}
\end{equation}

 The following equation determines the matrix factor m and n:
 \begin{equation}
\begin{cases}
    m=round\left(\frac{S-A}{S}\times3\right)\\
    m=n\\
    \end{cases}
\end{equation}

Where \(S\) denotes the area of entire frame while \(A\) is the sum of the area of the detected objects without overlap.

This determining method is based on a simple theory: for a particular frame, if there are enough objects or the object is big enough, human eyes are likely to ignore the distortion in the rest part of the frame. Otherwise, if there is on stare object or the objects themselves are small, human eyes are likely to notice heavy distortion in detail. 

Therefore, for frames where objects occupy a large area, we choose a relatively aggressive quantization matrix to achieve a higher compression rate. If the objects occupy only a small portion of the frame or no object is detected, we choose a relatively gentle quantization matrix.

\begin{equation}
    \begin{cases}
        I_{Y\_quantified(u,v)}=F_{Y\_quantified(u,v)} \times L_{m(u,v)}\\
        I_{U\_quantified(u,v)}=F_{U\_quantified(u,v)} \times C_{m(u,v)}\\
        I_{V\_quantified(u,v)}=F_{V\_quantified(u,v)} \times C_{m(u,v)}\\
    \end{cases}
\end{equation}


\subsection{The Theory}

The quantization process took the DCT matrix and divided each element by a corresponding value in the quantization matrix. A standard quantization matrix is shown below:

\begin{equation}
    \begin{bmatrix}
        16&11&10&16&24&40&51&61\\
        12&12&14&19&26&58&60&55\\
        14&13&16&24&40&57&69&56\\
        14&17&22&29&51&87&80&62\\
        18&22&37&56&68&109&103&77\\
        24&35&55&64&81&104&113&92\\
        49&64&78&87&103&121&120&101\\
        72&92&95&98&112&100&103&99\\
    \end{bmatrix}
\end{equation}

As indicated above, high-frequency values (values close to the bottom right corner) are divided by a relatively large number. With the divided values are rounded to the nearest integer value, most of these frequencies become 0 (dropped). The inverse quantization process will restore other frequencies to integer values close to their original values. 

When the image is again processed by JPEG encoder with a relatively high setting, the processed blocks will preserve large quantities of zeros in DCT form and reduce the data bits required for the encoding.


\section{Experiments}

\subsection{Experiment Setup}

\begin{itemize}

\item \textbf{Dataset: } Since this is an add-on to conventional encoders, and we use pre-trained YOLOv4 for the task, there is no training required for this method. (Retrain YOLOv4 is needed for this method to be applied to special occasions). For evaluation, we use the UVG dataset \cite{Mercat2020} as a source video clip to perform the experiment.

\item \textbf{Evaluation Method: }We use two methods to evaluate our method: quantitative analysis and sensory evaluation.
    \begin{itemize}
        \item The quantitative analysis uses MS-SSIM[33] to evaluate the overall distortion of our method. To measure the compression ratio boosted by our method, we use bits per pixel (Bpp) to denote the average data rate.
        
        \item The sensory evaluation computes the MS-SSIM between the original frame and the corresponding processed frame by pixel cloumns. The MS-SSIM of each pixel column is calculated and represented by a histogram style graph. 
    \end{itemize}
    
\item \textbf{Implementation Details: } 
The basic I/O operations and the color encoding system transform are done with OpenCV\cite{opencv_library}. To accelerate the process, we have implemented the rest of the procedure based on TensorFlow.
\bigbreak
The original clip is decoded and transformed by OpenCV then processed with a TensorFlow backend. The result is saved as a lossless PNG file to preserve the complete modifications. Later, the processed image sequence is sent to the FFmpeg program using a bitrate limiter to produce the final video clip. To better exclude uncontrollable variables, the original clip is also saved as a lossless PNG file separately.
\bigbreak
In the encoding phase, we have encoded the processed clip with the x264 encoder and the x265 encoder. The original image sequence was also encoded with x264 and x265 encoders directly. All the encoding procedures use two-pass encoding parameters to ensure the best encoding quality granted by the corresponding encoder. In total, we have encoded each sequence with six different bitrates: 400kbps, 800kbps, 1200kbps, 1600kbps, 2000kbps, and 2400kbps. Each corresponds to a Bpp of 0.2, 0.39, 0.58, 0.77, 0.96 and 1.16 respectively. In addition, ground truth is encoded with a lossless encoding parameter.
\bigbreak

Then, each encoded video clip was compared with the ground truth to calculate two types of parameters: MS-SSIM and PSNR. We have also isolated the rough object region and compare the result using the same metrics.

\end{itemize}
\subsection{Experiment Results}

\begin{figure}
    \centering
    \includegraphics[width=60mm]{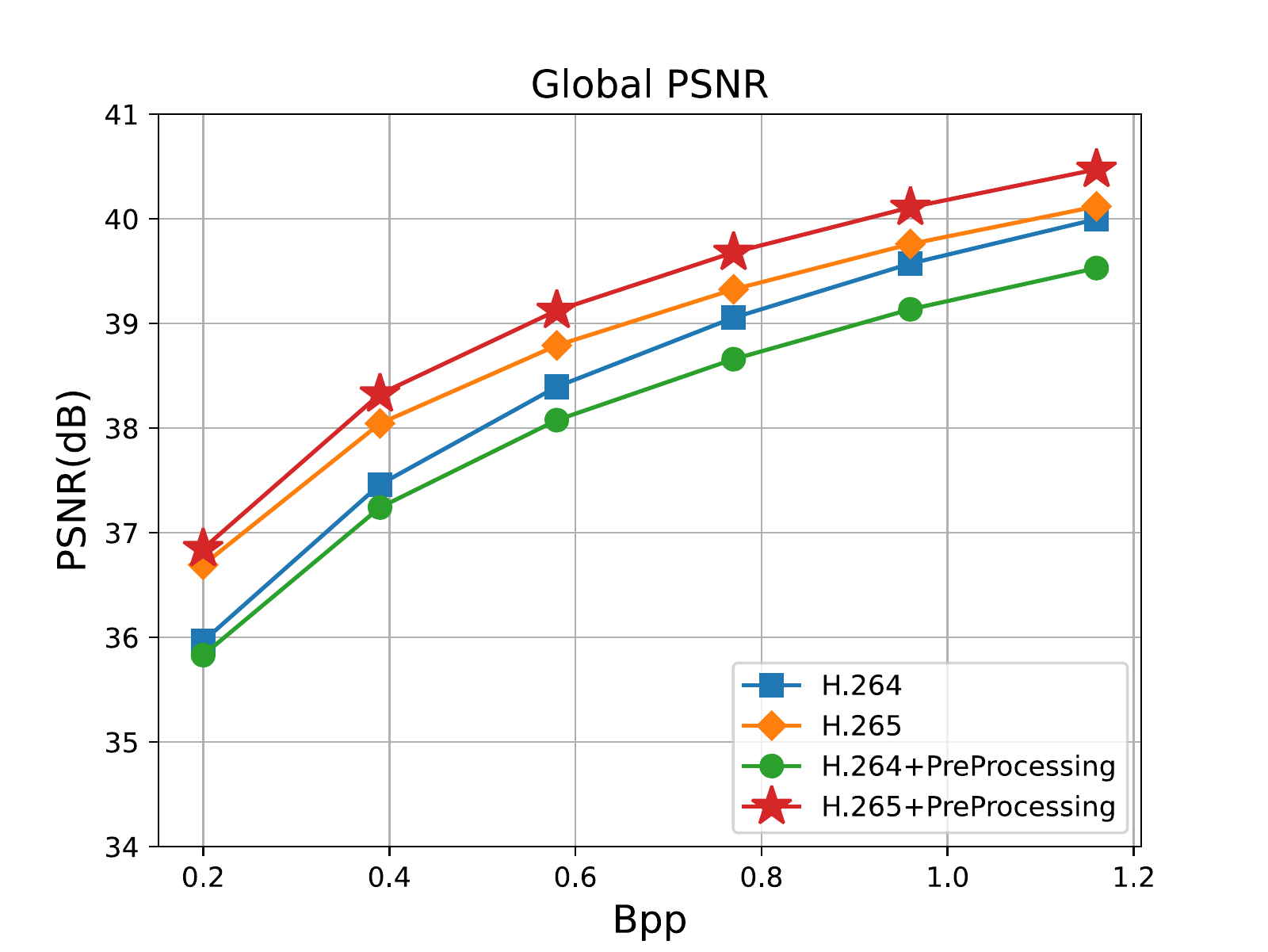}
    \includegraphics[width=60mm]{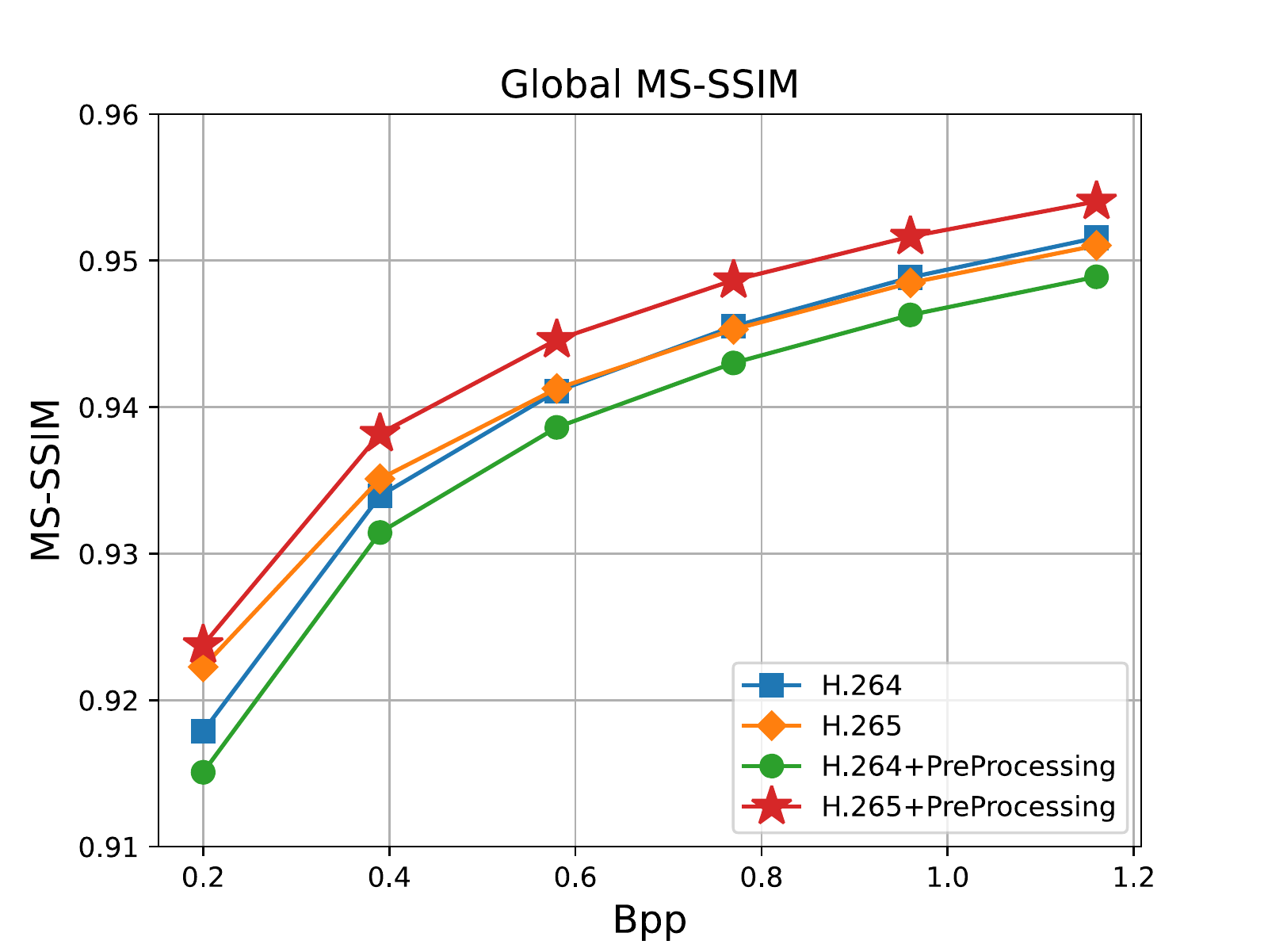}
    \caption{Comparison between the output of encoders with our preprocessing method and the output of direct encoding}
    \label{fig:GPSNR&GSSIM}
\end{figure}

In this section, we compared the MS-SSIM values and the PSNR values across different encoders with and without our preprocessing method. 

Firstly, we present the comparison between methods using PSNR and MS-SSIM metric. Each data point is the mean of all frames. Fig .~\ref{fig:GPSNR&GSSIM}  shows a promising result in which our preprocessing method works brilliantly with H.265 encoders. Benefit from the variable-size DCT, the H.265 encoders retains relatively high video quality with the help of our processing method. Both the PSNR and the MS-SSIM are greatly enhanced in the full-frame. Meanwhile, the H.264 encoders show a less promising result. The problem may lie in the quantization steps incompatible with our quantization matrix selection. 
\begin{figure}
    \centering
    \includegraphics[width=60mm]{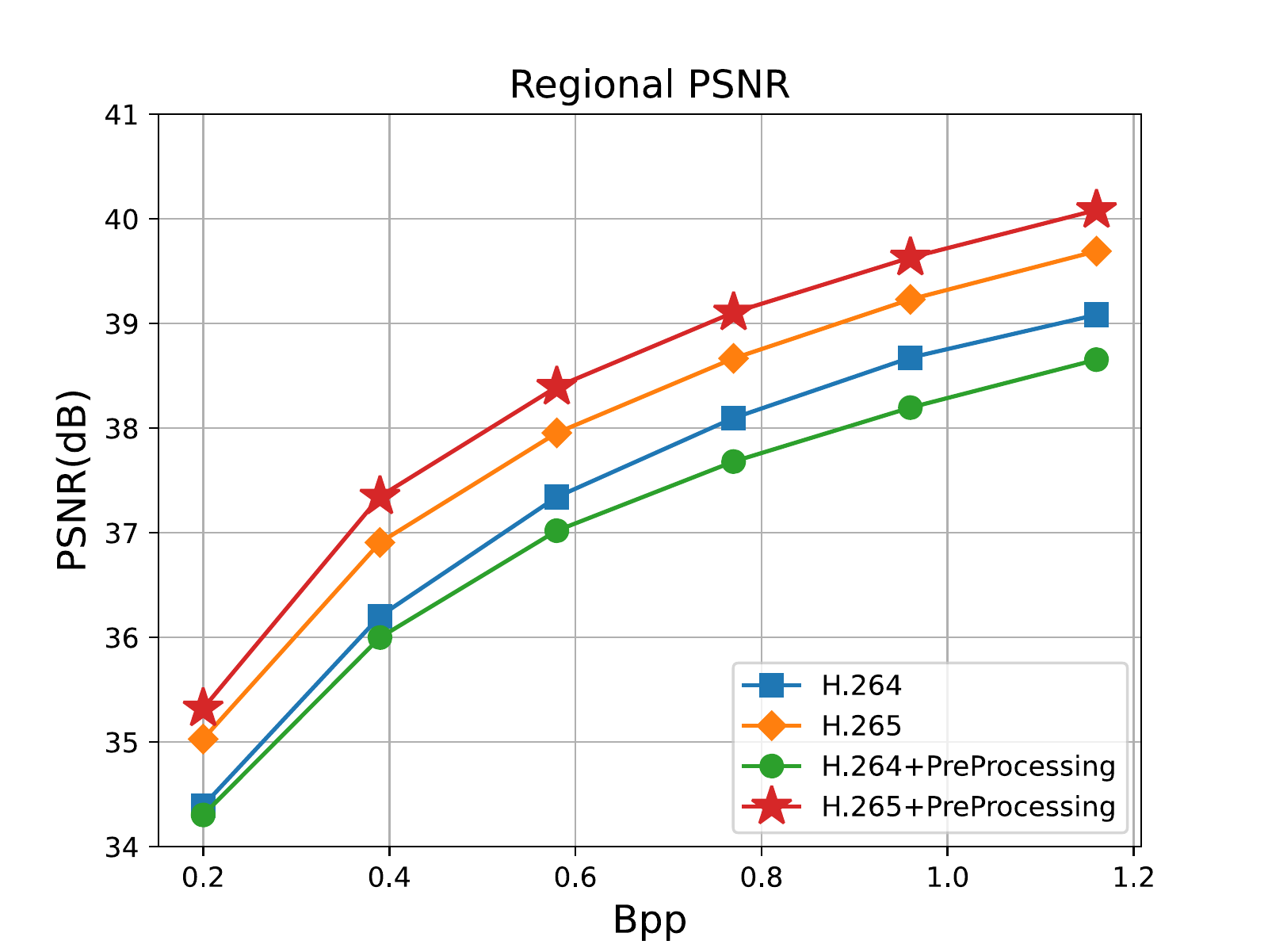}
    \includegraphics[width=60mm]{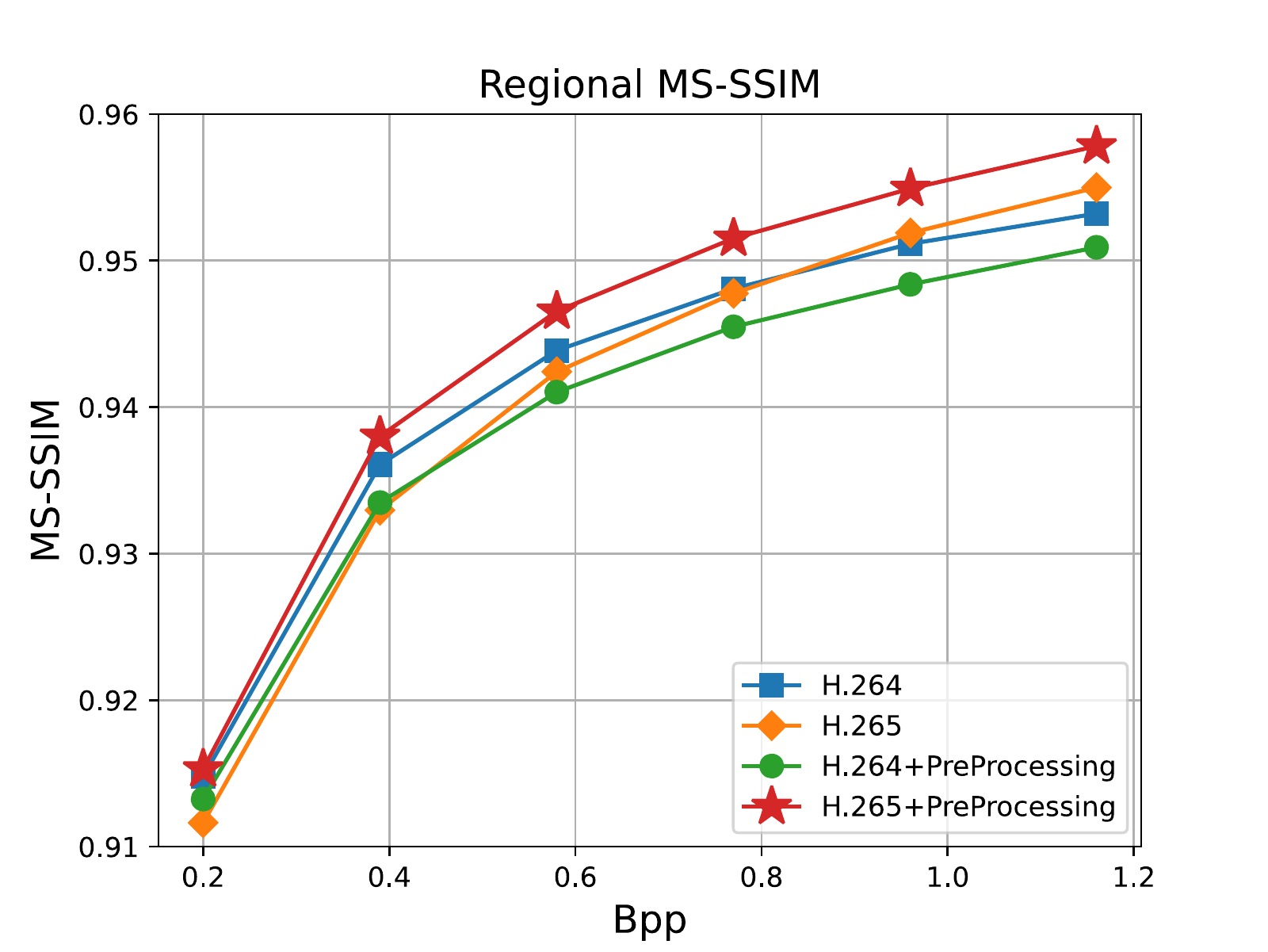}
    \caption{Comparison between the output of encoders with our preprocessing method and the output of direct encoding in Region of Interest}
    \label{fig:RPSNR&RSSIM}
\end{figure}

To better illustrate the effectiveness of our method, we also compared the MS-SSIM values and the PSNR values of different encoders in the object region. Fig .~\ref{fig:RPSNR&RSSIM} 
indicates that at the same Bpp, our method combining with H.265 encoder still maintain superior to other methods. One noticing point is that at an extremely low bitrate, our method combining with H.264 encoders performs better than direct encoding with H.265 encoders. This shows that in extreme conditions, our preprocessing method will help better preserve video quality.

\begin{figure}
    \centering
    \includegraphics[width=120mm]{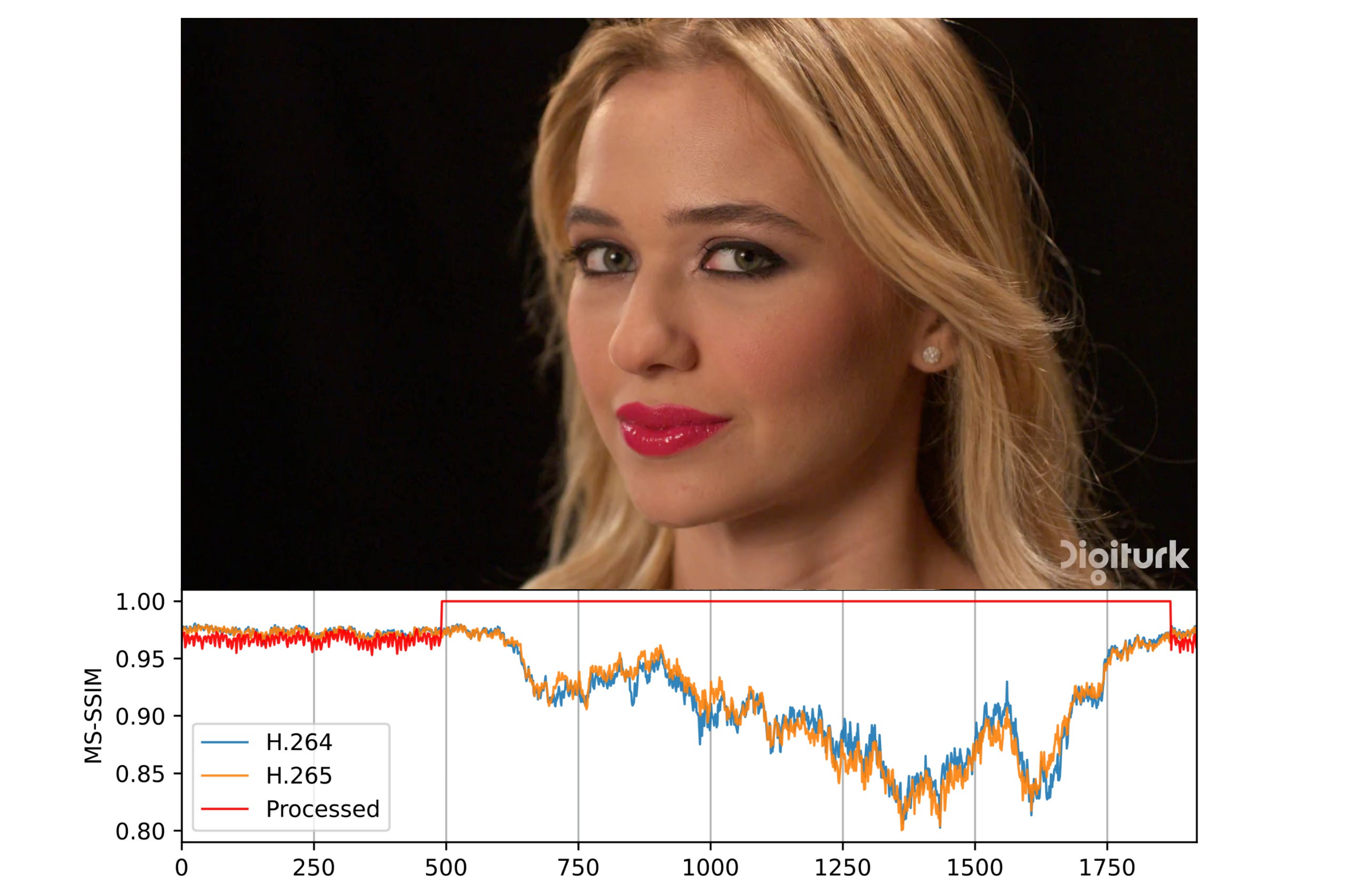}
    \caption{Column-wise MS-SSIM comparison between direct encoding and our output }
    \label{fig:composite}
\end{figure}

Finally, we compare the output of our method versus the direct output of the encoders using MS-SSIM values by column pixels. Note that the output file size was adjusted to roughly the same. 

Take frames like Fig .~\ref{fig:composite}, where the region of interests occupies a large area of the entire frame as an example. Most of us will concern about the details in the region of interest and ignore the detail loss in background. The graph area in the lower part of the figure shows that in horizontal pixel range [500, 1800], both H.264 and H.265 encoders have lost a significant amount of details in the region of interest while our method preserves the main body of the frame by reducing the details in the background. This method guarantees the viewing experience of the majority of the people to the greatest extent while reducing the bandwidth or the space needed for the video. Also, as present in the horizontal pixel range [0, 500] and [1800, 200] of the column-wise MS-SSIM figure, our method did not suffer from a significant loss in video quality in comparison with the direct encoding method.


\section{Conclusion}

In this paper, we have proposed a generic object detection based preprocessing method that can be applied before almost any DCT based encoder. Our method exploits the nature of DCT based encoding methods’ module and optimizes the input stream for encoding. The optimization is an adaptive process that is variable according to specific content. Substantially, our method improves the encoding result by telling the encoders, “what is important.” Furthermore, our method’s object detection is CNN based, which means that it can be deployed on machines with limited computing power and still able to perform the encoding process in a reasonable time.

Video clips are temporal data that always comes with audio information. Currently, our method considers video data as individual stills, and the determination of the quantization matrix is by the current frame only. Thus, the optimization is unaware of the temporal nature of the video. The next logical step is to perform sentiment analysis on audio data to further enhance the adaptive process. Moreover, the quantization matrices are still experience matrices for general purposes. With modern neural networks, these matrices’ size and value could be determined automatically based on the comprehension of the content.

\newpage
\bibliographystyle{splncs}
\bibliography{Formal}

\end{document}